# Diagnosis of Coronary Artery Disease Using Artificial Intelligence Based Decision Support System

Noor Akhmad Setiawan[1], P.A. Venkatachalam[2] and Ahmad Fadzil M.Hani[3]
Department of Electrical and Electronic Engineering, Universiti Teknologi PETRONAS,
Bandar Seri Iskandar, Tronoh 31750, Perak Darul Ridzuan, Malaysia
e-mail : [1] noorwewe@yahoo.com, [2] paruvachi_ammasai@petronas.com.my, [3] fadzmo@petronas.com.my

*Abstract*—This research is about the development a fuzzy decision support system for the diagnosis of coronary artery disease based on evidence. The coronary artery disease data sets taken from University California Irvine (UCI) are used. The knowledge base of fuzzy decision support system is taken by using rules extraction method based on Rough Set Theory. The rules then are selected and fuzzified based on information from discretization of numerical attributes. Fuzzy rules weight is proposed using the information from support of extracted rules. UCI heart disease data sets collected from U.S., Switzerland and Hungary, data from Ipoh Specialist Hospital Malaysia are used to verify the proposed system. The results show that the system is able to give the percentage of coronary artery blocking better than cardiologists and angiography. The results of the proposed system were verified and validated by three expert cardiologists and are considered to be more efficient and useful.

*Keywords*—Coronary artery disease, decision support system, diagnosis, fuzzy, rough set theory, reduct.

## I. Introduction

Artificial Intelligence (AI), including Artificial Neural Networks (ANN), fuzzy logic, evolutionary computing, machine learning and expert systems has been widely accepted as a tool for intelligent decision support system. Wide applications of AI in medical diagnosis are emerging in recent years. Nowadays, coronary artery disease is considered as the first killer disease in the world. [1][2]. Besides of that, the diagnosis of coronary artery disease is difficult, especially when there is no symptom. Much information from patients are needed in order to draw the correct diagnosis. It will be beneficial to use an advanced computer method such as artificial intelligence to build a decision support system for the diagnosis of coronary artery disease (CAD).

## II. Background

CAD is caused by the accumulation of plaques within the walls of the coronary arteries that supply blood to the myocardium. CAD may lead to continued temporary oxygen deprivation that will result in the damage of myocardium. The presence of CAD is considered to exist when the narrowing of at least one of the coronary arteries is more than 50%. Coronary angiogram or cardiac catheterization is considered as "gold standard" method to diagnose the presence of CAD. This method has high accuracy but it is invasive, expensive and not possible as a diagnosis for large population. Many research works have been conducted to diagnose the CAD using less expensive and non-invasive methods such as electrocardiogram (ECG) based analysis, heart sound analysis, medical image analysis, etc [2-5].

Development of computer methods for the diagnosis of heart disease attracts many researchers. At the earlier time, the use of computer is to build knowledge based decision support system which uses knowledge from medical experts and transfers this knowledge into computer algorithms manually. This process is time consuming and really depends on medical expert's opinion which may be subjective. To handle this problem, machine learning techniques have been developed to gain knowledge automatically from examples or raw data.

Detrano, et al, built a new discriminant function model for estimating probabilities of angiographic coronary disease [6]. This discrimination function operates based on logistic regression which is not interpretable easily. Modeling of heart disease using Bayesian network (also called belief network) is proposed by Jayanta and Marco [7][8]. Gamberger, et al, proposed Inductive Learning by Logic Minimization (ILLM). The aim of using machine learning technique is also to find the important and useful information extracted from medical data [9]. Another work is proposed by Yan, et al, by using multi layer perceptron to build decision support system for the diagnosis of five major heart diseases [10]. Research work on Rough Set Theory (RST) to model prognostic power of cardiac tests has been proposed by Komorowski and Ohrn. The work explores and identifies the need of a scintigraphic scan of a group of patients using rough set approach [11]. A research work on automated diagnosis on CAD based on rule induction and fuzzy modeling is proposed by Tsipouras, et al. The rule induction method that used to extract rules indirectly is C4.5 algorithm [12][13].

A decision support system for the diagnosis of CAD is proposed in this paper. RST based technique is used to discover the knowledge from CAD data sets in the form of decision rules. A hybrid RST and support based rule selection is also proposed.

The data set is taken from Data Mining Repository of University of California, Irvine (UCI)[14]. Finally the system is validated using data sets from Cleveland, Hungarian, Long Beach, Switzerland and from Ipoh Specialist Hospital, Malaysia.

## III. METHODOLOGY

### A. Data

CAD data sets from UCI are used in this work. The UCI-CAD data sets of 920 patients are collected from Cleveland Clinic Foundation, U.S., Hungarian Institute of Cardiology, Budapest, Hungary; Veterans Administration Medical Center, Long Beach, California, and University Hospital, Zurich, Switzerland. Most of researchers used. The attributes of these data sets are relating to physical examination, diagnostic laboratory and stress tests as shown in Table 1. The condition of CAD using these attributes is obtained using coronary angiography method as *num*. To build the decision support system, data sets of 661 selected patients from Cleveland, Hungarian and Long Beach are used. These selected patients have missing data in three input attributes which are *slope, ca* and *thal*. In this work, the missing data are imputed using ANN with RST based on complete data set [15-17]. We also use the data from 22 patients of Ipoh Specialist Hospital, Malaysia.

### B. Rule Generation

RST is used to discover the knowledge from the imputed data set. Data sets are represented in a tabular form as rows and columns. Each element of a row may represent an object or an instance which can be an event, or a patient, etc. Every column represents an attribute which can be a variable, an observation, a property, etc. that can be measured for each object. Such table can be called *information system*. In a formal manner, an information system is a pair $S = (U, A)$ where $U$ is a non-empty finite set of objects called the universe and $A$ is a non-empty finite set of attributes such that $\alpha : U \to V_\alpha$ for every $\alpha \in A$. The set $V_\alpha$ is called the value set of $\alpha$. In practical use of information system, there is an outcome of classification that is given as decision and expressed by single special attribute. This type of information system is called *decision system*. Formally, a decision system sometimes called decision table is an information system with the form $S = (U, A \cup D)$ where $D = \{d\}$ and $d \notin A$ is a decision attribute or simply decision. The elements of $A$ are called conditional attributes or simply conditions. A decision table is defined as:

$$DS = (U, C \cup D) \quad (1)$$

$D \not\subset C$ is called decision. According to Table 1, $D$ is attribute *num* (the presence or absence of CAD). $C$ is a set of conditions which are attributes *age, sex, …, ca* and *thal*. For $A \subseteq C$ and $x$ representing a patient, the indiscernibility relation can be defined as:

$$IND_C(A) = \{(x, x') \in U \times U \mid \forall c \in A, c(x) = c(x')\}. \quad (2)$$

The indiscernibility relation in equation (2) will induce a partition of $U$ into sets using only condition in $A$. Each object in the set cannot be distinguished from other object in the same set. The sets of classified objects are called equivalence classes denoted as $[x]_A$. Set approximation is used when a decision such as $d$ cannot be defined exactly.

For $A \subseteq C$, the approximations of $X \subseteq U$ using only information in $A$ are a lower-approximation $\underline{A}X$ and an upper-approximation $\overline{A}X$, which are defined as:

$$\underline{A}X = \{x \mid [x]_A \subseteq X\} \quad (3)$$

$$\overline{A}X = \{x \mid [x]_A \cap X \neq \emptyset\} \quad (4)$$

TABLE I
SUMMARY OF HEART DISEASE ATTRIBUTES

| Attribute | Description | Value description |
|---|---|---|
| age | Age | Numerical |
| sex | Sex | 1 if male; 0 if female |
| cp | Chest pain type | 1 typical angina<br>2 atypical angina<br>3 non-anginal pain<br>4 asymptomatic |
| trestbps | Resting systolic blood pressure on admission to the hospital (mmHg) | Numerical |
| chol | Serum cholesterol (mg/dl) | Numerical |
| fbs | Fasting blood sugar over 120 mg/dl ? | 1 if yes<br>0 if no |
| restecg | Resting electrocardiographic results : | 0 normal<br>1 having ST-T wave abnormality<br>2 LV hypertrophy |
| thalach | Maximum heart rate achieved | Numerical |
| exang | Exercise induced angina? | 1 if yes<br>0 if no |
| oldpeak | ST depression induced by exercise relative to rest | Numerical |
| slope | The slope of the peak exercise ST segment | 1 upsloping<br>2 flat<br>3 downsloping |
| ca | Number of major vessels colored by fluoroscopy | Numerical |
| thal | Exercise thallium scintigraphic defects | 3 normal<br>6 fixed defect<br>7 reversible defect |
| num | Diagnosis of heart disease (angiographic disease status / presence of coronary artery disease (CAD)) | 0 if less or equal than 50% diameter narrowing in any major vessel (CAD no)<br>1 if more than 50% (CAD yes) |

The set that contains objects that cannot be classified as precisely inside or outside $X$ is called boundary region (*BR*) of $X$:

$$BR_A(X) = \overline{A}X - \underline{A}X \quad (5)$$

$D$ is decision and $A \subseteq C$ as conditions, *A*-positive region of $D$ is defined as:

$$POS_A(D) = \bigcup_{X \in U/D} \underline{A}X \quad (6)$$

where $U/D$ represents the partition of $U$ according to decision $D$.

There are two issues in *DS*. Data set usually contains large part of objects with their attribute values. The first issue is that the same indiscernible objects may be represented more than one time. The second issue is that the attributes may be redundant or superfluous so that they can be reduced. The reduction of *DS* will result in reducts. A reduct is a minimal set of attributes $A \subset C$ such that $POS_A(D) = POS_C(D)$ amd is combination of conditions that can discern between objects as well as all conditions. Reducts can be computed using discernibility matrix and discernibility function [18].

Sometimes reducts computed based on objects relative discernibility are more interesting than full discernibility reduct, especially for decision rule generation. Once all of the relative reducts are determined, a set of decision rules can be generated from those reducts. Various algorithms are available to generate rules from reducts. RST can handle the discrete values only. Thus, discretization of numerical attributes must be done. Boolean reasoning algorithm is used to discretize the numerical attributes [18]. Number of "cuts" is generated during the discretization process.

Consider $DS = (U, C \cup D)$ as decision table. $\forall x \in U$ defines a series $c_1(x),...,c_k(x), d(x)$, where $\{c_1,...,c_k\} = C$ and $\{d\} = D$. Hence the decision rules can be generated in the form of $c_1(x),...,c_2(x) \to d(x)$. *C* can be the condition attributes of reduced form of decision table (reduct or relative reduct). This work uses relative reduct as a base to generate decision rules. The definition of *support* is described as how many objects match the corresponding rule. Support can be used as rule filtering criterion when there are too many rules generated.

*C. Rule Selection*

Many studies have been proposed for rule filtering methods [19-25]. In this paper hybrid RST based rule importance measure and support filtering to select the rules by converting the rules to decision tables is proposed. Filtering method based on rule support is applied to select the rules to reduce the number of rules before applying rule importance measure to select the most important rules. The modification is introduced by applying this method to decision system and converting rules to decision tables based on testing data which is complete CAD data set instead of training data which is incomplete CAD data set for rule importance measurement. Consider $R = \{Rule_1, Rule_2,..., Rule_j\}$ as a set of rules generated from training decision tables. If there are *i* objects on testing decision table, a new decision table $DS_{i \times (j+1)}$ can be formed.

The value of $Rule_a$ attribute of object $x_b$ is 1 if $Rule_a$ can be applied on $x_b$ both its antecedents and consequence. The value is 0 if the rule cannot be applied. The value equals decision value for column *j+1* with $a = 1,...,j$ and $b = 1,...,i$. The new decision table then can be reduced using RST reduct concept. The attributes of reduct are chosen as the selected rules based on their importance.

*D. Fuzzy Decision Support System (FDSS)*

The selected generated RST rules are crisp. Fuzzification must be applied to these crisp rules [26]. The fuzzy membership functions are in the form of triangular and trapezoid shapes. All numerical conditions are fuzzified based on the value of discretization "cuts". As an example, the numerical attribute *age* is discretized using Boolean reasoning into three discrete values which are [*,$\theta_1$), [$\theta_1$, $\theta_2$) and [$\theta_2$,*) which means "less than $\theta_1$", "equal or greater than $\theta_1$ and less than $\theta_2$ " and " equal or greater than $\theta_2$ " respectively. Therefore the "cuts" are $\theta_1$ and $\theta_2$. Two trapezoidal and single triangular membership functions of attribute *age* which are LOW, MEDIUM and HIGH can be generated as shown in Fig 1.

The same method is applied to the remaining numerical attributes where c can be any value. In this work, c is determined roughly. The nominal attributes have the crisp membership function. For simplification, no optimization is applied to the fuzzy membership functions.

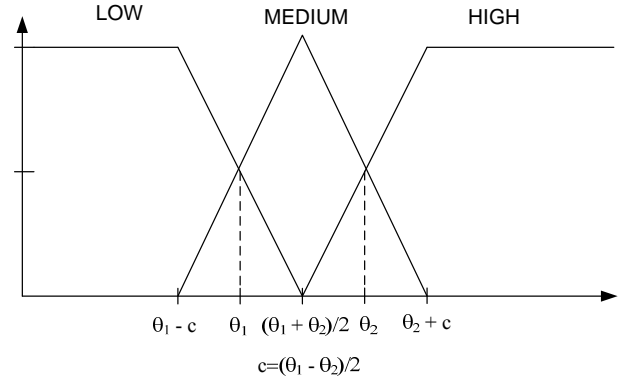

Fig. 1. Membership function of *age*

Fuzzy weighing method is proposed here based on support of selected RST rules from training data. If the *n*th crisp rule has support *sp(n)* then the corresponding fuzzy rule weight is:

$$w(n) = \frac{sp(n)}{\max(sp(1),...,sp(i))} \quad (7)$$

where *i* is the total number of rules.

The Mamdani inference engine is used for the inference process. Centroid defuzzification is chosen to get the numerical output of CAD diagnosis. For AND operator and implication, *min* function is used [27].

IV. RESULTS AND DISCUSSION

The system is developed using incomplete CAD data sets as the training set. This training set is imputed using ANN with RST. The imputed training set is then discretized. There are 358 objects (patients) in this training set. Using ROSETTA software, RST rule generation results in 3881 rules [18]. The proposed RST based rule selection method is able to select only 27 rules. Few rules out of 27 are shown in Table 2. This rule set is tested using complete CAD data set and compared to other methods as shown in Table 3 [25]. The selection method proposed in this work has better performance in accuracy and coverage.

TABLE II
SELECTED RST RULES

| Rule No. | Rules |
|---|---|
| 1. | oldpeak ([0.3,*)) AND slope (2) AND thal (7) => num (1) |
| 2. | fbs (0) AND thalach ([33,*)) AND slope ((1) AND ca ([*,1)) AND thal (3) => num (1) |
| 3. | fbs (0) AND cal([1,*)) AND thal (7) =>(1) |
| 4. | sex 1) AND fbs (0) AND thalach ([33,*)) AND excang (0) AND ca ([*,1)) AND thal (3) => num (1) |
| 5. | sex (1) AND fbs (0) AND restecg (0)) AND oldpeak ([0.3,*)) AND thal (7) => num (1) |
|  | ------------ ------------ ------------ |
| 27. | Age ([53,*)) AND tresbps ([129,*)) AND restecg (0) AND excang (0) AND ca ([*,1)) => num (1) |

TABLE III
RULE SELECTION PERFORMANCE

| Selection Methods | Accuracy | Coverage | Number of rules |
|---|---|---|---|
| Proposed Method | 0.852 | 0.937 | 27 |
| Support Based (Training Data) | 0.847 | 0.799 | 29 |
| Support Based (Testing Data) | 0.844 | 0.868 | 27 |
| Michalski μ=0.5 | 0.845 | 0.785 | 27 |
| Torgo | 0.845 | 0.785 | 27 |
| Brazdil | 0.845 | 0.785 | 27 |
| Pearson | 0.845 | 0.785 | 27 |
| Cohen | 0.863 | 0.65 | 29 |

To test the performance of fuzzy decision support system (FDSS), all four data sets from UCI-CAD is used which are Cleveland, Hungarian, Long Beach and Switzerland. All of the data sets contain missing values. Cleveland has only missing values in six objects. Switzerland has the largest number of missing values. For comparison, multi layer perceptron ANN, k-Nearest Neighbor, C4.5 and RIPPER method are implemented using WEKA software to diagnose the CAD on the four UCI-CAD data sets and Ipoh Specialist Hospital data set [28]. The results can be seen in Table 4 to 8. It can be seen that FDSS has good performance on all five data sets. Only k-NN method gives better performance on accuracy in Hungarian and Long Beach. This is because the training data is taken from these data sets. k-NN is based on similarity by computing Euclidean distance of the classified or testing objects to the training objects. The Hungarian and Long Beach have strong similarity to the training set. For Cleveland, Switzerland and Ipoh, FDSS gives better results.

TABLE IV
FDSS Performance on Cleveland Data Set

| Methods | Accuracy | Sensitivity | Specificity |
|---|---|---|---|
| FDSS | 0.83 | 0.81 | 0.85 |
| MLP-ANN | 0.81 | 0.77 | 0.85 |
| k-NN | 0.81 | 0.84 | 0.79 |
| C4.5 | 0.82 | 0.79 | 0.85 |
| RIPPER | 0.83 | 0.82 | 0.84 |

TABLE V
FDSS PERFORMANCE ON HUNGARIAN DATA SET

| Methods | Accuracy | Sensitivity | Specificity |
|---|---|---|---|
| FDSS | 0.84 | 0.70 | 0.91 |
| MLP-ANN | 0.54 | 0.44 | 1.00 |
| k-NN | 0.92 | 0.87 | 0.95 |
| C4.5 | 0.66 | 0.52 | 0.89 |
| RIPPER | 0.68 | 0.54 | 0.87 |

TABLE VI
FDSS PERFORMANCE ON LONG BEACH DATA SET

| Methods | Accuracy | Sensitivity | Specificity |
|---|---|---|---|
| FDSS | 0.75 | 0.83 | 0.49 |
| MLP-ANN | 0.76 | 0.75 | 1.00 |
| k-NN | 0.85 | 0.85 | 0.81 |
| C4.5 | 0.77 | 0.78 | 0.60 |
| RIPPER | 0.78 | 0.81 | 0.59 |

TABLE VII
FDSS PERFORMANCE ON SWITZERLAND DATA SET

| Methods | Accuracy | Sensitivity | Specificity |
|---|---|---|---|
| FDSS | 0.70 | 0.71 | 0.50 |
| MLP-ANN | 0.81 | 0.93 | 0.06 |
| k-NN | 0.62 | 0.96 | 0.10 |
| C4.5 | 0.53 | 0.97 | 0.10 |
| RIPPER | 0.41 | 0.96 | 0.08 |

TABLE VIII
FDSS PERFORMANCE ON IPOH DATA SET

| Methods | Accuracy | Sensitivity | Specificity |
|---|---|---|---|
| FDSS | 0.82 | 1.00 | 0.56 |
| MLP-ANN | 0.77 | 0.92 | 0.56 |
| k-NN | 0.73 | 0.69 | 0.78 |
| C4.5 | 0.55 | 0.85 | 0.11 |
| RIPPER | 0.68 | 0.69 | 0.67 |

FDSS is considered better than k-NN and MLP in term of knowledge transparency.

Comparison with three cardiologists on 30 selected patients from Cleveland data, the diagnosis of CAD by FDSS is better as shown in Table 9.

TABLE IX
Performance of FDSS and Three Cardiologists

| Metrics | Diagnosis Results | | | |
|---|---|---|---|---|
|  | FDSS | Cardiologist#1 | Cardiologist#2 | Cardiologist#3 |
| Accuracy | 0.87 | 0.67 | 0.67 | 0.73 |
| Sensitivity | 0.82 | 0.82 | 0.76 | 0.76 |
| Specificity | 0.92 | 0.46 | 0.54 | 0.69 |

FDSS is able to give somewhat actual percentage values of coronary artery blocking. But cardiologists or angiography cannot give such results. As an example, the results of blocking estimation by an expert cardiologist and FDSS on 30 patients from Cleveland are shown in Fig 2.

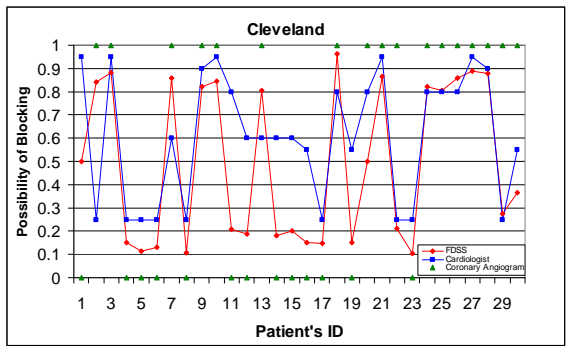

Fig. 2. Diagnosis result of FDSS, cardiologist and coronary angiography

## V. CONCLUSION

The development and performance of FDSS have been demonstrated in this paper. FDSS. The system is able to learn from the data sets and then discover the knowledge in the form of decision rules using RST. The knowledge is easy to understand because it is in the form of "IF-THEN" rules. The problem of having large number of rules is also solved using proposed RST based rule selection. The selected small number of rules still have good accuracy and coverage. These rules then fuzzified and used to build FDSS to diagnose the presence of CAD. FDSS is validated using UCI heart disease data sets collected from U.S., Switzerland and Hungary, and data from Ipoh Specialist Hospital Malaysia. Results are verified and validated by three cardiologists. The results show that FDSS is able to give the percentage of coronary artery blocking with better accuracy and considered to be more efficient.